\documentclass[runningheads]{llncs}
\usepackage{graphicx}
\usepackage{booktabs} % cz add
\usepackage{multirow} % cz add
\usepackage{bbm} % cz add
\usepackage{amssymb}
\usepackage{wrapfig} % cz add
\usepackage{array}

\usepackage{marvosym}
\usepackage[marginal]{footmisc}

\usepackage{bm}
\usepackage{amsmath}
\usepackage{hyperref}

\begin{document}
\setlength{\textfloatsep}{10pt}
\title{Personalized Retrogress-Resilient Framework for Real-World Medical Federated Learning}
\titlerunning{Personalized Retrogress-Resilient Federated Learning}
% \author{Paper ID 1293}
\author{Zhen Chen\textsuperscript{$*$}, Meilu Zhu\textsuperscript{$*$}, Chen Yang\textsuperscript{$*$}, Yixuan Yuan\textsuperscript{(\Letter)}}
%1{Zhen, Chen} %orcid{0000-0003-0255-6435}
%2{Meilu Zhu} %orcid{}
%3{Chen, Yang} %orcid{0000-0001-7841-5300}
%4{Yixaun Yuan} %orcid{0000-0002-0853-6948}

\authorrunning{Z. Chen, M. Zhu, C. Yang, and Y. Yuan.}
\institute{Department of Electrical Engineering,\\City University of Hong Kong, Kowloon, Hong Kong, China
	\email{yxyuan.ee@cityu.edu.hk}}
% \institute{ }
\maketitle              % typeset the header of the contribution
%
% {\tiny\footnote{$*$ Equal contribution.}}
\begin{abstract}
Nowadays, deep learning methods with large-scale datasets can produce clinically useful models for computer-aided diagnosis. However, the privacy and ethical concerns are increasingly critical, which make it difficult to collect large quantities of data from multiple institutions. Federated Learning (FL) provides a promising decentralized solution to train model collaboratively by exchanging client models instead of private data. However, the server aggregation of existing FL methods is observed to degrade the model performance in real-world medical FL setting, which is termed as \textit{retrogress}. To address this problem, we propose a personalized retrogress-resilient framework to produce a superior personalized model for each client. Specifically, we devise a Progressive Fourier Aggregation (PFA) at the server to achieve more stable and effective global knowledge gathering by integrating client models from low-frequency to high-frequency gradually. 
%Moreover, with a deputy model introduced to receive the aggregated server model, we design a Deputy-Enhanced Transfer (DET) at the client side to conduct three steps of \textit{Recover-Exchange-Sublimate} to ameliorate the personalized local model by transferring the global knowledge smoothly. 
Moreover, with an introduced deputy model to receive the aggregated server model, we design a Deputy-Enhanced Transfer (DET) strategy at the client and conduct three steps of \textit{Recover-Exchange-Sublimate} to ameliorate the personalized local model by transferring the global knowledge smoothly. 
Extensive experiments on real-world dermoscopic FL dataset prove that our personalized retrogress-resilient framework outperforms state-of-the-art FL methods, as well as the generalization on an out-of-distribution cohort. The code and dataset are available at \href{https://github.com/CityU-AIM-Group/PRR-FL}{https://github.com/CityU-AIM-Group/PRR-FL}.\footnote{$*$ Equal contribution.}
\end{abstract}
\section{Introduction}
Recent years have witnessed the superior performance of deep learning techniques in the field of computer-aided diagnosis~\cite{esteva2019guide,bhattacharya2020corrsignet,zhu2020alleviating}. By collecting large quantities of data from multiple institutions, tailored deep learning methods achieved great success and have been applied in the clinical practice to alleviate the workload of physicians \cite{liu2020clinical,esteva2017dermatologist,lotter2021robust}. 
\begin{figure}[thb]
    \centering
    \includegraphics[width=0.95\textwidth]{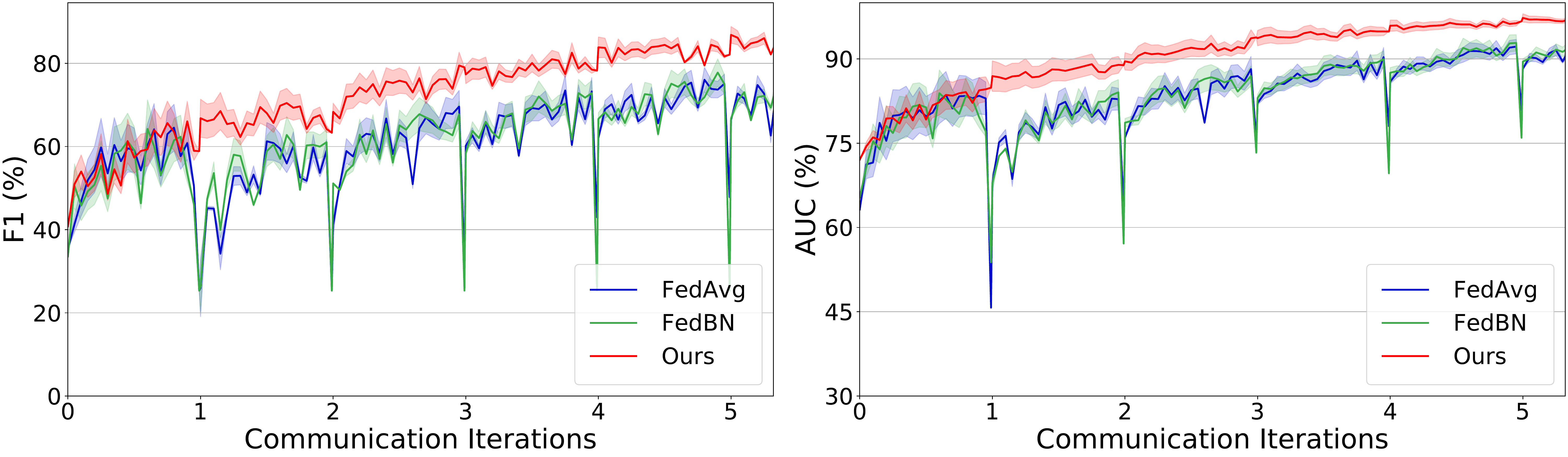}
    \caption{The training curves of client C in real-world dermoscopic FL dataset. The \textit{retrogress} happens in FedAvg \cite{fedavg} and FedBN \cite{li2021fedbn} after client-server communications. }
    \label{retrogress_curve}
\end{figure}
However, this is not a sustainable way to develop future intelligent healthcare systems, where the patient privacy and ethical concerns impede constructing centralized datasets with increasing size. To overcome this challenge, Federated Learning (FL) \cite{li2020federated,rieke2020future} provides a collaborative paradigm to optimize local models at multiple clients, and regularly aggregate these local models together at a global server. After that, the aggregated model is delivered to each client for the next local training. In this way, the client's data is kept private without being exposed to other participants, while the aggregated model with global knowledge can achieve superior performance, compared with the models trained at any single client. Among existing FL works, FedAvg \cite{fedavg,dou2021federated} proved that averaging local models trained at different clients is a generalization of updating the gradients of each batch when clients are assumed as independent and identically distributed (IID), which can reduce the communication overhead. To solve the practical non-IID distribution among clients, FedProx \cite{fedprox} introduced a proximal term to the objective of clients, regularizing the heterogeneous local updates to be closer to the initial global model.
%a regularization term of local model updates to restrain the convergence of FL training. 
Furthermore, FedBN \cite{li2021fedbn} and SiloBN \cite{andreux2020siloed} adopted personalized batch normalization (BN) settings for models at the client side to overcome the inter-client data heterogeneity. Since personalized FL allows each client to choose the aggregated server model or any intermediate client one during the local training \cite{federated_mutual,roth2020federated}, it is more suitable for medical FL scenarios.

In the scenarios of real-world medical FL, the inter-client data heterogeneity (e.g., caused by different imaging devices, protocols and even regional disease differences) is much more serious than aforementioned studies, thereby leading to enormous discrepancy between different local models. As illustrated in Fig.~\ref{retrogress_curve}, we observe existing FL methods encounter an abrupt performance drop (termed as \textit{retrogress}) after each communication between the server and clients. The aggregated models with \textit{retrogress} lose the previous knowledge and require re-adaptation to the client task during the local training, which hinders the local training of clients and the knowledge sharing of the server. We suppose the \textit{retrogress} in real-world FL may come from the following two reasons. On the one hand, when the severe data heterogeneity exists among clients, it is irrational to average these parameters in element-wise, because parameters from different clients may represent diverse semantic patterns at the same location~\cite{fedma}. In contrast, representing parameters in the frequency domain can ensure that the components are aligned along the frequency dimension, and also provide the flexibility to select the frequency band for aggregation \cite{frequency_pruning}. On the other hand, replacing the previous local models with the aggregated server model at the communication, obliterates the knowledge learned by the local models and degrades the optimization in the next iteration.

To address the \textit{retrogress} problem in real-world medical FL, we propose a personalized retrogress-resilient framework in both server and client perspectives, aiming to provide a customized model for each client. Specifically, we devise a Progressive Fourier Aggregation (PFA) to integrate client models at the server. Through Fast Fourier Transform (FFT) and inverse FFT (IFFT) to achieve the mutual conversion of client parameters to frequency domain, we average the low-frequency components of client parameters while preserving the remaining high-frequency components. By gradually increasing the frequency threshold of the shared components during the FL, PFA can effectively integrate the client knowledge in a manner consistent with network learning preferences. To the best of our knowledge, this work represents the first attempt to implement parameters aggregation of FL in frequency domain. At the client side, instead of replacing local models, we propose a Deputy-Enhanced Transfer (DET) to introduce a deputy model to receive the updated server model and maintain the personalized local model not contaminated. Considering that the deputy model suffers from the \textit{retrogress} after each communication, our DET conducts three steps, \textit{Recover-Exchange-Sublimate}, to recover the local prior and transfer the global knowledge to improve the personalized local model. Extensive experiments on real-world dermoscopic FL dataset demonstrate the effectiveness and out-of-distribution generalization of our personalized retrogress-resilient framework, which outperforms state-of-the-art FL methods.

\section{Personalized Retrogress-Resilient Framework}
\subsection{Overview}
\begin{figure}[t]
    \centering
    \includegraphics[width=0.80\textwidth]{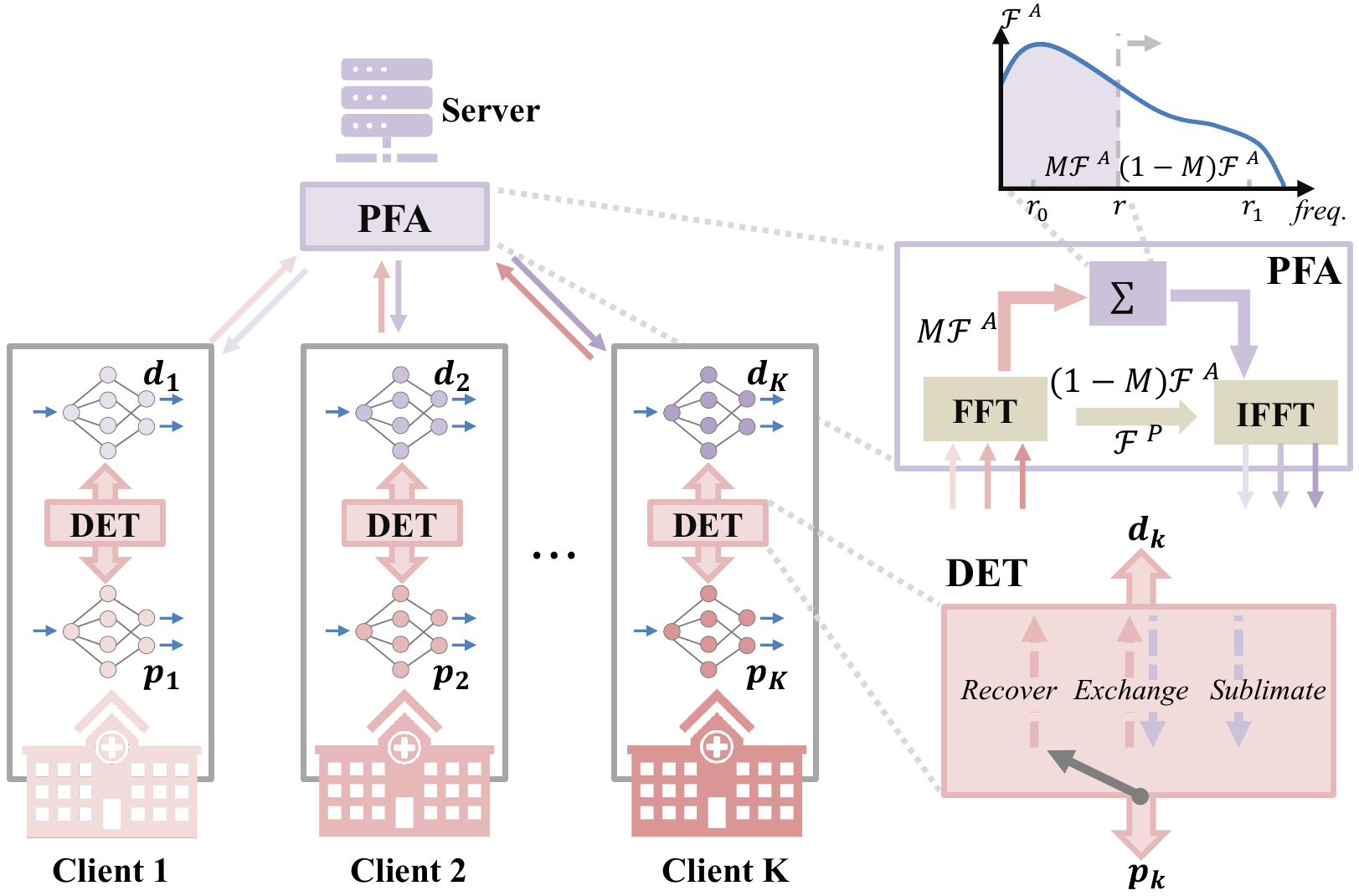}
    \caption{The personalized retrogress-resilient framework. At the server, Progressive Fourier Aggregation (PFA) integrates global knowledge in the frequency domain. At each client, Deputy-Enhanced Transfer (DET) is employed to improve the local personalized model without being disturbed by the communication.}
    \label{framework}
\end{figure}

Given a set of $K$ clients with their private data, our personalized retrogress-resilient framework aims to collaboratively generate a personalized model $\bm{p}$ with dominant performance for each client. These models $\{\bm{p}_{k}\}_{k=1}^{K}$ share the same network architecture to benefit from the server aggregation, which is the same as previous FL works. For the $k$-th client, the personalized model $\bm{p}_{k}$ is trained with private data for $E$ epochs locally, which is then uploaded to the server. The server collects client models, and aggregates them into server models with individual high-frequency components of client models using the proposed Progressive Fourier Aggregation (PFA). After that, these server models are delivered to corresponding client as a deputy model. The deputy model $\bm{d}_{k}$ can transfer the global knowledge through our Deputy-Enhanced Transfer (DET). Repeat these steps until local training reaches $T$ epochs. The overview of our personalized retrogress-resilient framework is illustrated in Fig.~\ref{framework}
%\vspace{-10pt}

\subsection{Progressive Fourier Aggregation in Server}
Previous FL works \cite{fedavg,fedprox,andreux2020siloed,li2021fedbn} gathered the global knowledge from different clients by directly element-wise averaging the parameters of local models to generate the aggregated server model. However, this coarse operation in parameter space violently degrades the model performance on clients, as the \textit{retrogress} of FedAvg \cite{fedavg} and FedBN \cite{li2021fedbn} observed in Fig.~\ref{retrogress_curve}. To alleviate the \textit{retrogress} induced by aggregation, we propose the Progressive Fourier Aggregation (PFA) to stably integrate client models in the frequency domain. 

Inspired by the fact that low-frequency components of parameters are the basis for the network capability \cite{frequency_pruning}, our PFA aggregates the relatively low-frequency components of parameters to share knowledge from different clients, while retraining their high-frequency components, which may contain specific knowledge for each individual client. Specifically, for a convolutional layer of $k$-th client model, we first reshape its parameter tensor $w_{k}\in \mathbb{R}^{N\times C\times d_{1}\times d_{2}}$ into a 2-D matrix $w_{k}^{'}\in \mathbb{R}^{d_{1}N\times d_{2}C}$, where $N$ and $C$ represent the output and input channels, and $d_{1}$ and $d_{2}$ are the spatial shape of the kernel. Then, we obtain the amplitude map $\mathcal{F}^{A}$ and phase map $\mathcal{F}^{P}$ through the Fourier transform $\mathcal{F}=\mathcal{F}^{A}e^{j \mathcal{F}^{P}}$, as follows:
\begin{equation}
    \mathcal{F}(w_{k}^{'})(m, n)=\sum_{x, y} w_{k}^{'}(x, y) e^{-j 2 \pi\left(\frac{x}{d_{1}N} m+\frac{y}{d_{2}C} n\right)}, j^{2}=-1,
\end{equation}
which can be implemented efficiently using the FFT~\cite{fft}. To extract the low-frequency components for aggregation, we employ a low-frequency mask $M$ with zero values except for the central region:
\begin{equation}
    {M}(m, n)=\mathbbm{1}_{(m, n) \in[-rd_{1}N: rd_{1}N,-rd_{2}C: rd_{2}C]},
\end{equation}
where $r\in(0,0.5)$ represents the low-frequency threshold, and the center of $w_{k}^{'}$ is set as the coordinate $(0, 0)$. By averaging the low-frequency components over clients, the aggregated frequency components for $k$-th client are calculated as:
\setlength{\abovedisplayskip}{3pt}
\setlength{\belowdisplayskip}{3pt}
\begin{equation}
    \hat{\mathcal{F}}^{A}(w_{k}^{'})=(1-M) \circ \mathcal{F}^{A}(w_{k}^{'})+\frac{1}{K}\sum_{k=1}^{K}M \circ \mathcal{F}^{A}(w_{k}^{'}),
\end{equation}
where $\circ$ is the element-wise multiplication. Considering the fact that networks are trained to learn the low-frequency knowledge prior to the high-frequency counterpart~\cite{image_prior,xu2019training,rahaman2018spectral}, we design a progressive strategy to implement our PFA by increasing $r=r_{0}+\frac{r_{1}-r_{0}}{T}t$ during the FL training, where $r_{0}$ and $r_{1}$ are the initial and terminated low-frequency threshold. 

Finally, after applying the inverse Fourier transform $\mathcal{F}^{-1}$ to convert the amplitude and phase maps back to parameters as $\hat{w}_{k}^{'}=\mathcal{F}^{-1}([\hat{\mathcal{F}}^{A}(w_{k}^{'}), \mathcal{F}^{P}(w_{k}^{'})])$, we obtain the aggregated parameters of $k$-th client. The same PFA can be applied to the fully connected layers of client models, without the reshape of parameters.
%\vspace{-20pt}
\subsection{Deputy-Enhanced Transfer in Client}
Our PFA can alleviate the \textit{retrogress} induced by aggregation at the server, however, directly updating client models with the aggregated server parameters still obliterates the learned local knowledge and further degrades the optimization in the next iteration. To solve this problem, we propose the Deputy-Enhanced Transfer (DET) to subtly fuse the global knowledge with the local prior, rather than the direct replacement. 
In addition to a personalized local model $\bm{p}$, each client also contains a deputy model $\bm{d}$ to receive the aggregated parameters from server. The proposed DET conducts three steps: \textit{Recover}, \textit{Exchange} and \textit{Sublimate}, to smoothly transfer the global knowledge from the deputy $\bm{d}$ to the personalized local model $\bm{p}$.

\noindent\textit{\textbf{Recover}}. When the deputy $\bm{d}$ is updated with the aggregated model $\bm{s}$ from the server, its performance encounters a severe deterioration due to the aforementioned \textit{retrogress} problem. Therefore, we firstly regard the personalized local model $\bm{p}$ as a teacher to recover the deputy model $\bm{d}$ with the local knowledge. In this step, the personalized local model $\bm{p}$ is supervised by the cross entropy loss, while the deputy model $\bm{d}$ is optimized by a joint loss function $L_{\bm{d}}$:
\begin{equation}
\label{equation:det:1}
    L_{\bm{d}} = L_{CE1} + \sum_{i=1}^{N} {p} ({x}_i)\log \frac{{p} ({x}_i) }{{q} ({x}_i)},
\end{equation}
where ${x}_{i}$ is the $i$-th training sample and $N$ is the number of training samples on the current client. $L_{CE1}$ is the cross entropy loss. The ${p}({x}_i)$ and ${q} ({x}_i)$ are the posterior probabilities of the deputy $\bm{d}$ and personalized local model ${\bm{p}}$. The second term of Eq.~\eqref{equation:det:1} is the Kullback-Leibler (KL) divergence that helps the deputy $\bm{d}$ to quickly recover the adaptability to the client with performance improved. This step is crucial to guarantee the deputy model does not hurt the personalized local model in the next step.

\noindent\textit{\textbf{Exchange}}. 
Once the recovered performance of the deputy $\bm{d}$ is close to the teacher $\bm{p}$, as $\phi_{val}(\bm{d})\geq \lambda_1\phi_{val}(\bm{p})$, where $\phi_{val}$ represents a specific performance metric (\textit{e.g.}, F1) on the validation set, we conduct the mutual learning \cite{deep_mutual} between the deputy $\bm{d}$ and the personalized local model $\bm{p}$ to exchange the global knowledge and the local knowledge. Here, the deputy model $\bm{d}$ is supervised by Eq.~\eqref{equation:det:1} and the personalized local model $\bm{p}$ is learned by the following loss function $L_{\bm{p}}$:
\begin{equation}\label{equation:det:2}
    L_{\bm{p}} = L_{CE2} + \sum_{i=1}^{N} \textit{q} (\textit{x}_i)\log \frac{\textit{q} (\textit{x}_i) }{\textit{p} (\textit{x}_i)}.
\end{equation}
The second term of Eq.~\eqref{equation:det:2} transfers the global knowledge of the server to the personalized local model $\bm{p}$ through the deputy $\bm{d}$.
In this way, the knowledge exchange can promote the generalization ability of all clients~\cite{deep_mutual}.

\noindent\textit{\textbf{Sublimate}}. Finally, when the performance of the deputy $\bm{d}$ is highly close to the personalized local model $\bm{p}$, as $\phi_{val}(\bm{d}) \geq \lambda_2\phi_{val}(\bm{p})$, where $0<\lambda_{1}<\lambda_{2}<1$, the deputy model $\bm{d}$ serves as the teacher to further guide $\bm{p}$ with $L_{\bm{p}}$ in Eq.~\eqref{equation:det:2}, which enables the global knowledge can be transferred to the personalized local model to the greatest extent.

%\vspace{-5pt}
\begin{table}[b]
	\centering
\renewcommand\arraystretch{1.0}
\setlength{\tabcolsep}{1.5pt}	
	\caption{Detailed information of real-world dermoscopic FL dataset.}\label{fl_dataset}
    \begin{tabular}{p{1cm}<{\centering}|p{2cm}<{\centering}|p{2.5cm}<{\centering}|p{2cm}<{\centering}|p{2cm}<{\centering}}
        \toprule[1pt]
        Client         & Nevus & Benign Keratosis & Melanoma & Total \\ \hline
        A & $1832$  & $475$              & $680$      & $2987$  \\
        B & $3720$  & $124$              & $24$       & $3868$  \\
        C & $803$   & $490$              & $342$      & $1635$  \\
        D & $1372$  & $254$              & $374$      & $2000$  \\
        \bottomrule[1pt]
    \end{tabular}
\label{tab0}
\end{table}

\begin{table}[thb]
\centering
	\caption{Comparison with state-of-the-art on real-world dermoscopic FL dataset.}\label{sota_compare}
\renewcommand\arraystretch{1.0}
\setlength{\tabcolsep}{1.5pt}	
\begin{tabular}{c|cccc|c|cccc|c}
    \toprule[1pt]
    \multirow{2}{*}{Method} & \multicolumn{5}{c|}{F1 (\%)} & \multicolumn{5}{c}{AUC (\%)} \\ \cline{2-11}
            & A  & B  & C  & D & Avg & A    & B    & C    & D   & Avg  \\
            \hline
    FedAvg \cite{fedavg}  &  $57.44$  &  $48.14$  &  $56.80$  & $44.20$  &  $51.64$   &   $81.52$   &   $82.75$   &   $76.14$   &  $71.14$   &   $77.89$      \\
    FedProx \cite{fedprox} &  $56.70$  &  $39.09$  &  $54.70$  & $45.95$  &   $49.11$  &   $81.70$   &   $70.09$   &   $76.76$   &  $74.83$   &   $75.84$      \\
    SiloBN \cite{andreux2020siloed} &  $50.83$  &  $63.81$  &  $53.98$  & $61.90$  &   $57.63$  &   $83.17$   &   $81.41$   &   $77.90$   &   $80.56$  &   $80.76$     \\
    IDA \cite{yeganeh2020inverse}    &  $55.62$  &  $41.87$  &  $55.42$  & $45.64$  &   $49.64$  &   $81.27$   &   $75.95$   &   $78.38$   &  $73.10$   &   $77.18$       \\
    FedBN \cite{li2021fedbn}  &  $54.96$  &  $72.10$  &  $54.73$  &  $62.07$ &  $60.96$   &   $83.06$   &   $96.35$   &   $79.97$   &  $81.36$   &  $85.18$      \\
    FML \cite{federated_mutual}  &  $69.14$  &  $75.83$  &  $66.02$  & $59.40$  &  $67.60$   &   $88.38$   &   $95.49$   &   $85.05$   &   $82.81$  &   $87.93$     \\
    \hline
    Ours \textit{w/o} DET        &  $67.28$  &  $75.20$  &  $61.44$  &  $59.24$ &  $65.79$   &   $87.95$   &   $89.81$   &   $83.76$   &   $81.39$  &   $85.73$   \\
    Ours \textit{w/o} PFA        &  $69.41$  &  $77.90$  &  $66.77$  & $61.90$  &  $69.00$   &   $88.70$   &   $97.03$   &   $84.25$   &  $82.59$   &   $88.14$     \\
    Ours        &  $\bm{72.00}$ &  $\bm{79.70}$  &  $\bm{68.87}$  &  $\bm{62.46}$ &  $\bm{70.75}$   &   $\bm{88.98}$   &   $\bm{97.58}$   &   $\bm{85.22}$   &  $\bm{83.90}$   &   $\bm{88.92}$      \\
    \bottomrule[1pt]
\end{tabular}
\end{table}
\section{Experiment}
\subsection{Dataset and Implementation Details}
\textbf{Real-World Dermoscopic FL Dataset}.
Considering that the real-world FL in medical scenarios is involved with complex data heterogeneity, we construct a novel FL benchmark dataset of dermoscopic images to evaluate various FL methods under the actual client settings. Specifically, we collect $8,940$ and $2,000$ images that are diagnosed as nevus, benign keratosis or melanoma from HAM10K \cite{ham10k} and MSK \cite{msk} dataset, respectively. To preserve the critical shape information of irregular-sized images for lesion diagnosis, we first crop the central square regions using the shorter side length, and then unify them into $128\times 128$ resolution. With the metadata of each image in HAM10K \cite{ham10k}, we segregate the images with the same institution and imaging device into one client. Consequently, our FL dataset contains 4 clients, including 3 clients from HAM10K \cite{ham10k} and MSK \cite{msk} data as one. Detailed information of the dataset is illustrated in Table~\ref{fl_dataset}. We further divide samples of each client into the training, validation and test sets as 7:1:2, ensuring the consistent proportion of categories among these three sets. In this way, our dataset can reflect the difficulty of practical FL deployment, including the data heterogeneity, different imaging devices and various sample numbers among clients.

\noindent \textbf{Implementation Details}.
We implement our personalized retrogress-resilient framework and state-of-the-art FL methods using VGG-16BN \cite{vgg} in PyTorch \cite{pytorch}. During the local training at the client side, the networks are optimized using SGD with the batch size of $16$. The learning rate is initialized as $1\times 10^{-2}$ and halved after every $25$ epochs. The communication is conducted after every $E=5$ epochs in local training until reaching $T=250$ epochs in total. The frequency thresholds $r_{0}$ and $r_{1}$ in PFA are set to $0.35$ and $0.48$, and the performance thresholds $\lambda_1$ and $\lambda_2$ in DET are set to $0.7$ and $0.9$, respectively. We choose F1 score as the metric $\phi_{val}$ to measure the performance of deputy and personalized models. The best model is selected using the hold-out validation set on each client. The training employs a combination of random flip and rotation as data augmentation to avoid overfitting. All experiments are performed on NVIDIA V100 GPU. 

\noindent \textbf{Evaluation Metrics}.
The F1 and AUC are utilized to measure the performance of various FL methods comprehensively. For the 3-category classification, we calculate these two metrics of each category separately, and then perform the macro average to eliminate the impacts of imbalance. The higher scores of both F1 and AUC indicate better performance on the dermoscopic diagnosis task. In addition to showing the performance of FL methods on each client, we also calculate the marco average of 4 clients for a more concise comparison.
%\vspace{-10pt}
%\subsection{Experimental Results and Analysis}
\subsection{Comparison with State-of-the-art Methods}
To evaluate the performance of our personalized retrogress-resilient framework, we perform a comprehensive comparison with the state-of-the-art FL methods, including FedAvg \cite{fedavg}, FedProx \cite{fedprox}, SiloBN \cite{andreux2020siloed}, IDA \cite{yeganeh2020inverse}, FedBN \cite{li2021fedbn} and FML \cite{federated_mutual}. As illustrated in Table \ref{sota_compare}, our method achieves the best performance, with the overwhelming average F1 of $70.75\%$ and average AUC of $88.92\%$. 
Noticeably, our framework outperforms the personalized FL method, FedBN~\cite{li2021fedbn}, by a large margin, \textit{e.g.}, $9.79\%$ in average F1 and $3.74\%$ in average AUC. This advantage confirms that the proposed PFA and DET can alleviate the \textit{retrogress} in FedBN~\cite{li2021fedbn}.
Moreover, compared with FML \cite{federated_mutual} that also employs the deputy model at each client, our method obtains superior performance with a remarkable increase of $3.15\%$ in average F1 and $0.99\%$ in average AUC. These experimental results demonstrate the performance advantage of our method over state-of-the-art FL methods in the scenarios of real-world medical FL. 

\begin{table}[tb]
\centering
\caption{Generalization comparison of different FL works on out-of-distribution data.}
\centering
\renewcommand\arraystretch{1.1}
\setlength{\tabcolsep}{2.9pt}
\begin{tabular}{c|c c c c c c|c}
    \toprule[1pt]
    \multirow{1}{*}{Method} & \multicolumn{1}{c}{FedAvg\cite{fedavg}} & \multicolumn{1}{c}{FedProx\cite{fedprox}}& \multicolumn{1}{c}{SiloBN\cite{andreux2020siloed}}& \multicolumn{1}{c}{IDA\cite{yeganeh2020inverse}}& \multicolumn{1}{c}{FedBN\cite{li2021fedbn}}& \multicolumn{1}{c|}{FML\cite{federated_mutual}}& \multicolumn{1}{c}{Ours}\\ 
    \hline
    F1 (\%)  & $37.76$ & $38.14$ &  $42.87$ & $40.11$ & $43.74$ &  $37.01$ & $\bm{50.20}$ \\
    AUC (\%) & $66.49$ &  $67.36$ & $65.65$ & $62.75$ &  $70.85$ &  $66.81$ & $\bm{73.70}$ \\         
    \bottomrule[1pt]
\end{tabular}
\label{out-of-distribution}
\end{table}
%\vspace{-30pt}

\subsection{Ablation Study}
To further validate the effectiveness of PFA and DET, we perform the detailed ablation study in Table \ref{sota_compare}. Specifically, we implement two ablative baselines of the proposed personalized retrogress-resilient framework, by removing the DET (denoted as \textit{w/o} DET) and the PFA (denoted as \textit{w/o} PFA) individually. As illustrated in Table \ref{sota_compare}, the proposed PFA and DET can bring a F1 improvement of $1.75\%$ and $4.96\%$, respectively. Compared with FedBN \cite{li2021fedbn}, our method \textit{w/o} DET only replaces the parameter aggregation with PFA, which proves that the PFA results in a $4.83\%$ F1 increase. Moreover, when applying the same parameter aggregation with a deputy at each client, our method \textit{w/o} PFA outperforms FML \cite{federated_mutual} with a $1.40\%$ increase in F1, which can be attributed to the effect of the tailor-made DET with \textit{Recover-Exchange-Sublimate} three steps. The ablation experiment confirms that the proposed PFA and DET play an important role in solving data heterogeneity in real-world medical FL scenarios, resulting in the performance advantage of our personalized retrogress-resilient framework.

\subsection{Out-of-distribution Generalization}
To further verify the generalization ability on the out-of-distribution cohort, we test the trained models on an unseen cohort with $427$ samples from another institution \cite{ham10k}.
As shown in Table \ref{out-of-distribution}, our method achieves the best generalization with F1 of $50.20\%$ and AUC of $73.70\%$ in such challenging scenario. Moreover, our method demonstrates superior classification performance in comparison to existing algorithms \cite{fedavg,fedprox,andreux2020siloed,yeganeh2020inverse,li2021fedbn,federated_mutual} with statistically significant increments of $12.44\%$, $12.06\%$, $7.33\%$, $10.09\%$, $6.46\%$ and $13.19\%$ in average F1 score. These competitive experimental results can indicate the superiority of our method on the generalization ability. 
%\vspace{-10pt}
\section{Conclusion}
In this work, the \textit{retrogress} problem is observed in existing FL methods after every aggregation.
%, which is induced by parameters aggregation at server and parameters updating at client. 
To promote the FL in real-world medical scenarios, we propose a personalized retrogress-resilient framework with modification in both the server and clients.  Specifically, we conduct PFA to integrate the global knowledge from low-frequency to high-frequency gradually at the server.
At the client side, instead of replacing local models, DET 
conducts three steps of \textit{Recover-Exchange-Sublimate} to transfer global knowledge to improve the personalized model without being disturbed by \textit{retrogress}.
Extensive experiments confirm the advantages of our framework over state-of-the-art works on a real-world dermoscopic FL dataset, as well as the generalization on an out-of-distribution cohort. 

\subsubsection{Acknowledgments.} This work is supported by Shenzhen-Hong Kong Innovation Circle Category D Project SGDX2019081623300177 (CityU 9240008) and CityU SRG 7005229.

\bibliographystyle{splncs04}
\bibliography{ref}




% LNCS Homepage, \url{http://www.springer.com/lncs}. Last accessed 4
% \end{thebibliography}
\end{document}